\documentclass[letterpaper, conference]{IEEEtran}
\IEEEoverridecommandlockouts

\ifCLASSOPTIONcompsoc
  \usepackage[nocompress]{cite}
\else
  \usepackage{cite}
\fi
\ifCLASSINFOpdf
\else
\fi

\usepackage{amssymb}
\usepackage{multirow}
\usepackage{multicol}
\usepackage{url}
\usepackage{array}
\usepackage{verbatim}
\usepackage{float}
\usepackage{units}
\usepackage{textcomp}
\usepackage{amsmath}
\usepackage[utf8]{inputenc}
\usepackage{graphicx}
\usepackage{caption}
\usepackage{subcaption}
\usepackage{setspace}
\usepackage{placeins}
\usepackage{rotating}
\usepackage{mdwmath}
\usepackage{varwidth}
\usepackage{algpseudocode}

\def\BibTeX{{\rm B\kern-.05em{\sc i\kern-.025em b}\kern-.08em
    T\kern-.1667em\lower.7ex\hbox{E}\kern-.125emX}}

\newcommand{\MYfooter}{\smash{
\hfil\parbox[t][\height][t]{\textwidth}{\centering
\thepage}\hfil\hbox{}}}

\makeatletter

\def\ps@headings{%
\def\@oddhead{\parbox[t][\height][t]{\textwidth}{\centering
XXXXXXXXXXXXXXXXXXXXXXXXXXXXXXXXXXXXXXX\\
}\hfil\hbox{}}%

\def\@evenhead{\parbox[t][\height][t]{\textwidth}{\centering
XXXXXXXXXXXXXXXXXXXXXXXXXXXXXXXXXXXXX\\
}\hfil\hbox{}}%

\def\@oddfoot{\MYfooter}%
\def\@evenfoot{\MYfooter}}

\def\ps@IEEEtitlepagestyle{%
\def\@oddhead{\parbox[t][\height][t]{\textwidth}{\centering
}\hfil\hbox{}}%
\def\@evenhead{\scriptsize\thepage \hfil \leftmark\mbox{}}%
\def\@oddfoot{ XXX-X-XXXX-XXXX-X/XX/\$31.00 \textcopyright 2023 IEEE \hfil 
\leftmark\mbox{}}%
\def\@evenfoot{\MYfooter}}

\makeatother
\begin{document}
%

\title{A Hybrid ConvNeXt-EfficientNet AI Solution for Precise Falcon Disease Detection}

\author{\IEEEauthorblockN{Alavikunhu Panthakkan$^{1, a}$, Zubair Medammal$^{2}$, S M Anzar$^{3}$, Fatma Taher $^{4}$ and Hussain Al-Ahmad$^{5}$}
\IEEEauthorblockA{\textit{$^{1 \& 5}$College of Engineering and IT, University of Dubai, U.A.E.}\\
\textit{$^{2}$Department of Zoology, University of Calicut, Kerala, India}\\
\textit{$^{3}$Department of Electronics and Communication, TKM College of Engineering, Kollam, India}\\
\textit{$^{4}$4NextGen Center Academic Director, Zayed University, U.A.E}\\
\textit{Corresponding Author: $^{a}$apanthakkan@ud.ac.ae}
} \\}


%


\maketitle
\IEEEpubid{\begin{minipage}{\textwidth}\ \\ \\[12pt] 
\normalsize 979-8-3503-4524-7/23/\$31.00
\copyright2023 IEEE. 
\end{minipage}} 
\begin{abstract}
Falconry, a revered tradition involving the training and hunting with falcons, requires meticulous health surveillance to ensure the health and safety of these prized birds, particularly in hunting scenarios. This paper presents an innovative method employing a hybrid of ConvNeXt and EfficientNet AI models for the classification of falcon diseases. The study focuses on accurately identifying three conditions: Normal, Liver Disease and `Aspergillosis'. A substantial dataset was utilized for training and validating the model, with an emphasis on key performance metrics such as accuracy, precision, recall, and F1-score. Extensive testing and analysis have shown that our concatenated AI model outperforms traditional diagnostic methods and individual model architectures. The successful implementation of this hybrid AI model marks a significant step forward in precise falcon disease detection and paves the way for future developments in AI-powered avian healthcare solutions.
\end{abstract}

\IEEEpeerreviewmaketitle
\section{Introduction}
In the intricate domain of falconry, an art steeped in historical significance, the health and well-being of falcons are of utmost importance. These birds, highly valued in various cultural and environmental settings, require advanced and precise health management techniques, especially during rigorous activities like hunting~\cite{fischer2014comparison}. In response to this need, our paper presents a pioneering approach in avian healthcare, leveraging the advancements in artificial intelligence.

This research amalgamates the powerful capabilities of two neural network architectures, ConvNeXt and EfficientNet~\cite{panthakkan2023unleashing}, creating a hybrid AI model that excels in the detection and classification of falcon diseases. This hybrid model not only captures the robust feature extraction of ConvNeXt but also benefits from the efficient scaling characteristics of EfficientNet, making it a formidable tool in the realm of veterinary diagnostics.

Our study is anchored on a comprehensive dataset that encompasses a wide range of falcon health conditions, focusing primarily on the identification of Normal', Liver Disease', and `Aspergillosis' cases. The model's performance was rigorously evaluated using key metrics such as accuracy, precision, recall, and F1-score. The results demonstrate that this concatenated AI model significantly surpasses traditional diagnostic methods and individual architectures in terms of accuracy and reliability.

The successful implementation of this hybrid AI model in falcon disease detection marks a significant advancement in veterinary science. It not only enhances the precision in diagnosing falcon ailments but also lays the groundwork for future innovations in AI-powered solutions for avian healthcare, contributing to the conservation and understanding of these magnificent birds.
    \begin{figure}[t]
	\begin{center}
	\includegraphics[scale=0.275]{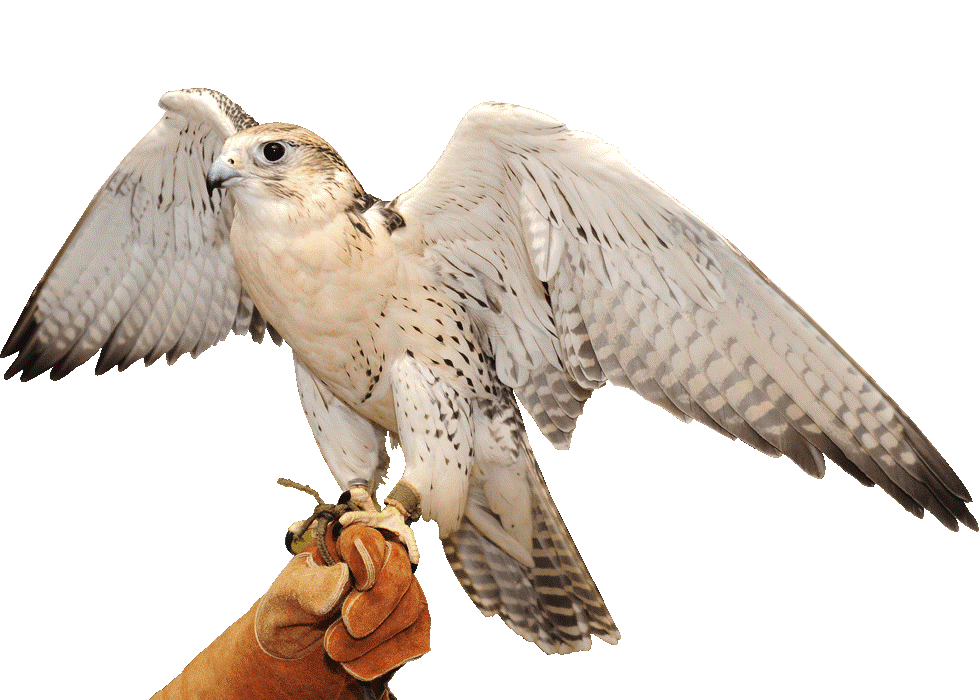}
		\caption{UAE Falcon}
		\label{fig:Cancatenated_Binar_IoT}
	\end{center}
    \end{figure}
\section{Background Work}
Falcons are a group of birds of prey belonging to the genus Falco, known for their remarkable speed, keen eyesight, and exceptional hunting skills. Ranging widely in size, these birds are characterized by their tapered wings and agile flight, enabling them to dive at high speeds to catch prey mid-air. Falcons inhabit a diverse array of environments across the world, from Arctic tundras to tropical forests, adapting remarkably to different ecological settings~\cite{muller2006study}.

Historically and culturally, falcons hold a significant place in many societies. Falconry, the art of training falcons for hunting, dates back thousands of years and is revered in various cultures for its symbiosis between human and bird. In some regions, falcons are not only hunting companions but also symbols of status and heritage~\cite{fischer2014comparison}.

Biologically, falcons are fascinating creatures. They possess a unique anatomy that allows for their incredible flight speeds and hunting prowess. For instance, their sharp, hooked beaks and strong talons are adapted for capturing and killing prey. They also have a specialized bone structure in their nostrils, which helps in breathing at high speeds and altitudes~\cite{hemeiri2021retrospective}.

In the context of conservation, several falcon species have faced challenges due to habitat loss, pesticide use, and other human-induced factors. Efforts in conservation and rehabilitation have been pivotal in the resurgence of some species, such as the Peregrine Falcon, which was once on the brink of extinction in parts of its range.

The health of falcons is a critical aspect, especially in falconry and conservation efforts. Diseases can significantly impact their population and well-being~\cite{alrefaei2020molecular}. Hence, advancements in disease detection and health management, such as the AI solutions discussed here are crucial for the ongoing conservation and care of these remarkable birds.

\subsection{Falcon Diseases} 
The section on Falcon Diseases delves into the various health issues that can adversely affect these birds of prey, highlighting the significance of vigilant monitoring and early intervention in falconry practices. Two primary diseases discussed are 'Liver' disease and 'Aspergillosis,' each posing unique threats to the health and well-being of falcons.

Liver Disease in falcons is a serious condition that can stem from nutritional imbalances or infectious agents. Symptoms may include an enlarged liver (hepatomegaly), jaundice, and changes in behavior. This disease can significantly impact a falcon's health, affecting vital functions like metabolism, detoxification, and digestion. Falcons suffering from liver disease often experience decreased energy levels, weakened immune function, and impaired hunting or flying abilities. Early diagnosis and appropriate treatment are crucial in managing this condition and mitigating its effects on the falcon's overall health.

Aspergillosis in falcons is a respiratory disease caused by inhaling spores of the Aspergillus fungus, a common environmental presence. This inhalation can lead to respiratory infections, with symptoms such as breathing difficulties, lethargy, reduced appetite, and nasal discharge~\cite{tarello2011etiologic}. The impact of Aspergillosis can be severe, leading to chronic respiratory issues and impaired flight capabilities. In advanced stages, the disease can be fatal, as the fungus primarily targets the respiratory tract, causing inflammation and reducing respiratory efficiency. Early detection and treatment are vital in managing Aspergillosis, with veterinarians and falconers playing key roles in monitoring health, implementing preventive measures, and providing necessary medical interventions.

\subsection{Literature Survey}
The literature on falcon disease detection and management has evolved significantly, with various studies contributing to a deeper understanding of avian health. Research from the Abu Dhabi Falcon Hospital, as detailed by Müller et al.~\cite{muller2006study}, has identified E.coli septicemia as a prevalent condition in falcons, often linked to feeding problems, particularly in overtrained or exhausted birds. This condition has been observed to be more prevalent in younger falcons but can affect birds up to six years of age, with the potential to cause fatal intoxication.

Another crucial area of research, as explored by Müller et al.~\cite{muller2008outbreak}, focuses on diagnosing Enterocytozoon Bieneusi infections in falcons. This study highlights the potential risk these infections pose, not only to the birds themselves but also to falconers, drawing attention to the zoonotic aspect of falcon diseases.

The significance of radiography, endoscopy, and haematological profiling in falcon medicine has been underscored in the work of Mohamed et al.~\cite{mohamed2014radiographic}. These diagnostic tools have been identified as indispensable in accurately diagnosing and treating a range of conditions in falcons.

A retrospective study by Hemeiri et al.~\cite{hemeiri2021retrospective} brings to light the primary health issues faced by falcons, including ingluvitis, aspergillosis, and bacterial enteritis. This study emphasizes the risk of transmission of these diseases from falcons to their owners, underlining the need for regular health check-ups and efficient disease control.

In broader avian health research, Sadeghi et al.~\cite{sadeghi2023early} investigated the application of thermography and machine learning in detecting Avian Influenza and Newcastle Disease in broilers. The study concluded that this combination proves to be a valuable tool for early disease detection, significantly reducing economic losses in poultry production.

The integration of machine learning (ML) and deep learning (DL) into avian medicine, particularly in falconry, has marked a significant advancement in veterinary practices. Initially, ML applications in avian disease detection were limited to statistical analysis and basic pattern recognition, facing challenges in handling unstructured data such as images or complex physiological signals. However, the emergence of DL, especially convolutional neural networks (CNNs), has revolutionized medical imaging and data interpretation in falcon disease detection. Advanced DL architectures like ConvNeXt and EfficientNet have shown remarkable efficiency in processing detailed features in avian anatomy and pathology, thus enhancing the accuracy of diagnosing complex diseases in falcons.

The practical applications of DL in falcon healthcare have extended beyond diagnostic imaging to include the analysis of clinical symptoms and behaviors, improving both diagnostic accuracy and the overall welfare of the birds. Despite these advancements, challenges related to data availability, quality, and the interpretability of DL models, as well as ethical considerations in the use of AI in veterinary medicine, persist.

Looking to the future, the integration of DL models with wearable technologies for real-time health monitoring presents promising prospects for proactive falcon healthcare. Such advancements could lead to more preventative care approaches and extend the benefits of these technologies to other avian species, potentially revolutionizing veterinary care across a broad spectrum of bird species. The ongoing research and development in ML and DL applications in falcon disease detection are significant, offering substantial potential in enhancing disease diagnosis and treatment. However, these advancements necessitate careful consideration of data quality, model interpretability, and ethical implications, promising not only to transform falcon health management but also to provide valuable insights applicable to wider avian healthcare.
    \begin{figure*}[ht]
	\begin{center}
	\includegraphics[scale=0.175]{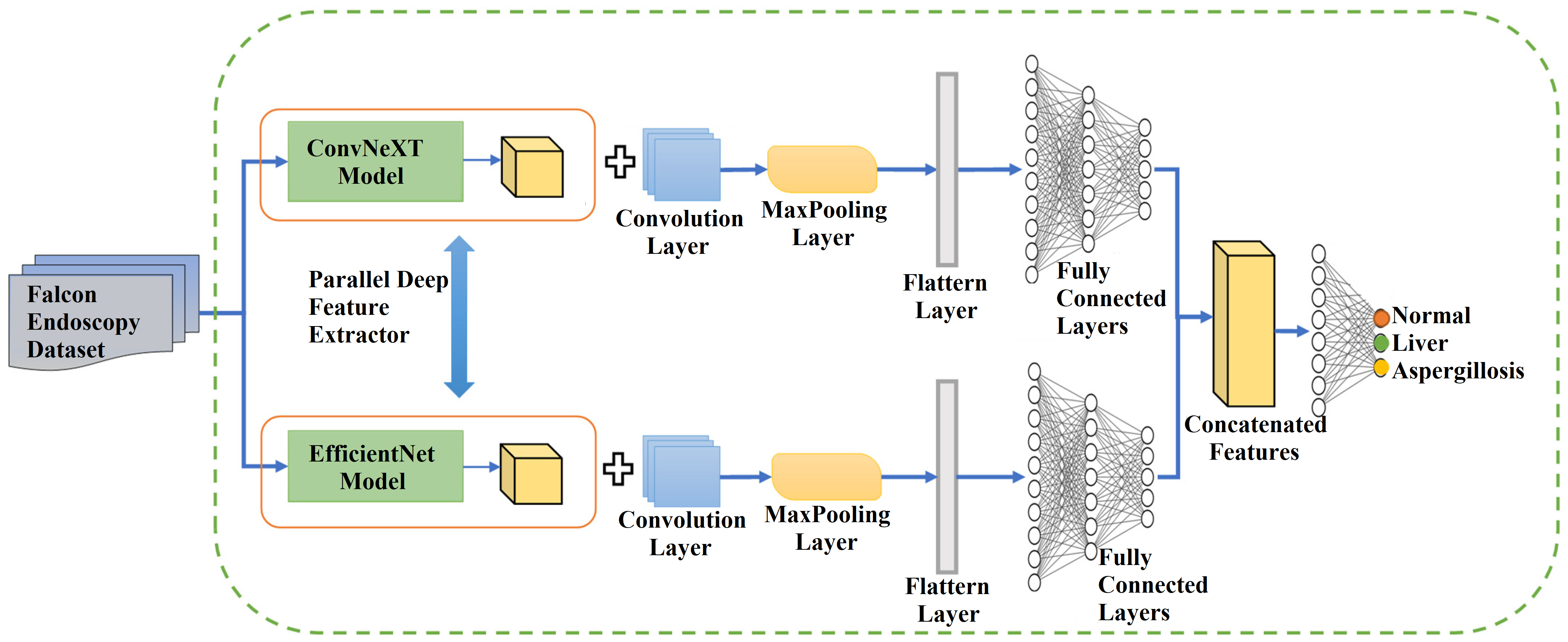}
		\caption{Block Diagram of the Proposed Concatenated ML Model}
		\label{fig:Cancatenated_Binar_IoT}
	\end{center}
    \end{figure*}
\section{Proposed Concatenated ML Model}
The advent of deep learning models has been a transformative force in artificial intelligence, especially in applications requiring the processing and interpretation of complex data sets, such as in precise falcon disease detection. These models excel in advanced pattern recognition, which is essential for identifying the subtle signs of disease in falcons, often surpassing the capabilities of traditional diagnostic methods. Additionally, their ability to handle multifaceted data, including images, sounds, and physiological metrics, makes them particularly suitable for analyzing the varied aspects of falcon health.

Among deep learning models, the ConvNeXt and EfficientNet models stand out for their specific contributions to falcon disease detection. The ConvNeXt model, an evolution in the field of convolutional neural networks, is tailored for image recognition and classification. It proves invaluable in analyzing visual data, such as scans and camera images, to detect signs of disease that are not discernible to the naked eye. Its architecture facilitates deeper and more intricate pattern recognition, which is crucial in identifying early stages of diseases in falcons.

On the other hand, the EfficientNet model is renowned for its scalability and efficiency. This model is designed to process various scales of data effectively, maintaining high accuracy without necessitating a proportional increase in computational resources. Its versatility allows it to handle diverse data inputs, from high-resolution images to detailed physiological data, making it an ideal choice for medical diagnostic applications.

The methodology employed in our study begins with the meticulous assembly of a comprehensive dataset, crucial for the effectiveness of the AI model. This dataset is composed of endoscopy images that illustrate a variety of falcon health conditions, including but not limited to 'Normal,' 'Liver' disease, and 'Aspergillosis.' The richness and diversity of this dataset are fundamental to the model's training process, as it provides a wide spectrum of data for the system to learn from.

Leveraging transfer learning techniques, the model undergoes an efficient training process on this curated dataset. Transfer learning is a powerful approach in machine learning, allowing the model to apply knowledge gained from one task to another related task. This technique is particularly useful in fine-tuning the model to the specific nuances of falcon diseases.

Once the initial training phase is complete, the hybrid architecture, combining the ConvNeXt and EfficientNet models, is meticulously fine-tuned. This fine-tuning process is critical in honing the model's ability to detect and interpret subtle patterns within the data, which are indicative of specific falcon diseases. The ConvNeXt model contributes its profound visual analysis capabilities to this hybrid solution, enabling it to recognize intricate visual details in the endoscopy images. In tandem, the EfficientNet model brings its scalable and efficient data processing strengths to the table, ensuring that the system can handle large datasets without compromising on speed or accuracy.

The culmination of these efforts is a robust, accurate, and efficient AI solution for diagnosing diseases in falcons. This hybrid AI system harnesses the best of both worlds: the deep, detailed visual analysis of ConvNeXt, and the scalable, resource-efficient processing of EfficientNet. The synergy of these models results in a groundbreaking advancement in the field of falcon disease detection, significantly enhancing the precision and effectiveness of diagnoses. This development not only marks a notable achievement in veterinary AI but also contributes immensely to the ongoing efforts in health management and conservation of these majestic birds.
\begin{figure*}[t!]
\begin{subfigure}{0.49\linewidth} 
\includegraphics[scale=0.6]{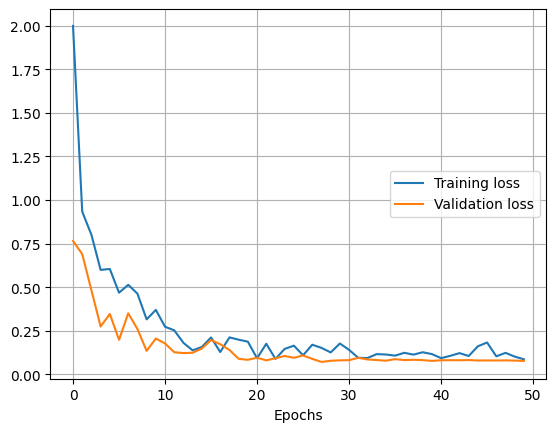}
\caption{Training and Validation Loss Versus Epoch}
\label{fig:CM-Decision Tree Model}
\end{subfigure}
\begin{subfigure}{0.49\linewidth} 
\includegraphics[scale=0.6]{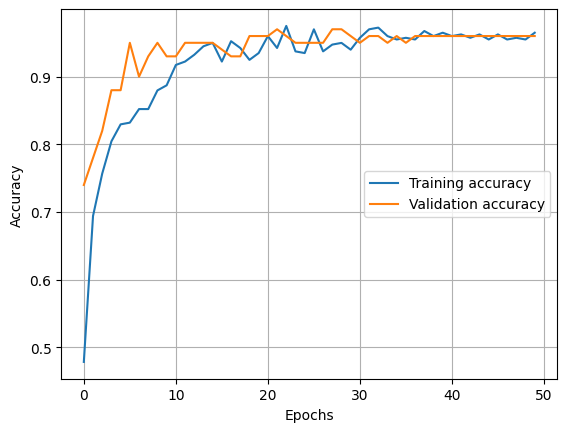}
\caption{Training and Validation Accuracy Versus Epoch}
\label{fig:Test}
\end{subfigure}
\caption{Training Performance of the Proposed Model} \label{fig:Train_Test}
\end{figure*}
\begin{figure*}[t!]
\begin{subfigure}{0.49\linewidth} 
\includegraphics[scale=0.7]{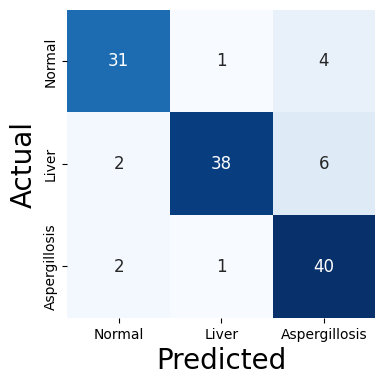}
\caption{ConvNeXt Model}
\label{fig:CM-Decision Tree Model}
\end{subfigure}
\begin{subfigure}{0.49\linewidth} 
\includegraphics[scale=0.7]{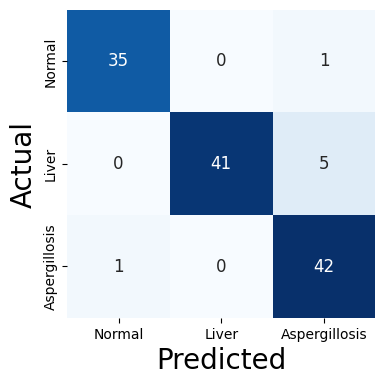}
\caption{EfficientNet Model}
\label{fig:Test}
\end{subfigure}
\caption{Confusion Matrix of the Base Models} \label{fig:Conf1}
\end{figure*}
    \begin{figure}[ht]
	\begin{center}
	\includegraphics[scale=0.7]{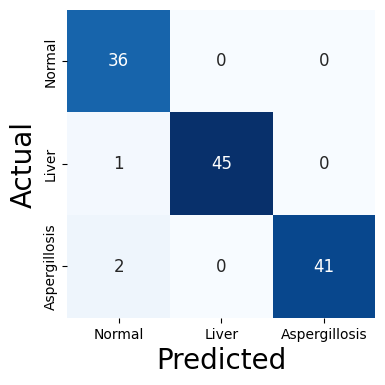}
		\caption{Confusion Matrix of the Proposed Concatenated Model}
		\label{fig:Conf2}
	\end{center}
    \end{figure}
    \begin{figure}[t]
	\begin{center}
	\includegraphics[scale=0.6]{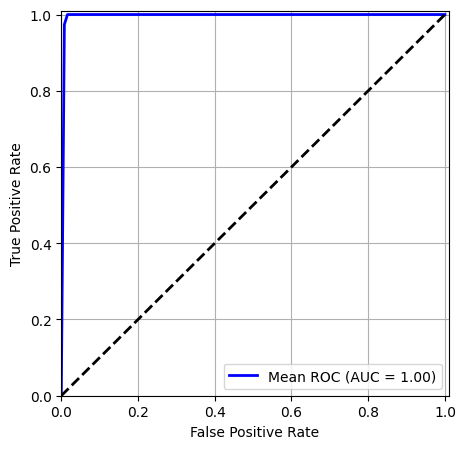}
		\caption{ROC-AUC Plot of the Proposed Concatenated Model}
		\label{fig:ROC}
	\end{center}
    \end{figure}
\begin{table*}[t!]
\begin{center}
\caption{Performance Evaluation of the Baseline and Proposed Model}
\label{tab:Table1}
\scalebox{1.0}{
\begin{tabular}{llllll}
\hline
\multicolumn{1}{c}{AI Models} & \multicolumn{1}{c}{Types} & \multicolumn{1}{c}{Accuracy} & \multicolumn{1}{c}{Precision} & \multicolumn{1}{c}{Recall} & \multicolumn{1}{c}{F1-Score} \\ 
\hline
 & \multicolumn{1}{c}{Normal} & \multicolumn{1}{c}{0.87} & \multicolumn{1}{c}{0.89} & \multicolumn{1}{c}{0.86} & \multicolumn{1}{c}{0.87} \\ 
 & \multicolumn{1}{c}{Liver} & \multicolumn{1}{c}{0.89} & \multicolumn{1}{c}{0.95} & \multicolumn{1}{c}{0.83} & \multicolumn{1}{c}{0.88} \\ 
\multicolumn{1}{c}{ConvNeXt} & \multicolumn{1}{c}{Aspergillosis} & \multicolumn{1}{c}{0.86} & \multicolumn{1}{c}{0.80} & \multicolumn{1}{c}{0.93} & \multicolumn{1}{c}{0.86} \\ 
 & \multicolumn{1}{c}{Average} & \multicolumn{1}{c}{0.87} & \multicolumn{1}{c}{0.88} & \multicolumn{1}{c}{0.87} & \multicolumn{1}{c}{0.87} \\ 
\hline
 & \multicolumn{1}{c}{Normal} & \multicolumn{1}{c}{0.97} & \multicolumn{1}{c}{0.97} & \multicolumn{1}{c}{0.97} & \multicolumn{1}{c}{0.97} \\ 
 & \multicolumn{1}{c}{Liver} & \multicolumn{1}{c}{0.95} & \multicolumn{1}{c}{1.00} & \multicolumn{1}{c}{0.89} & \multicolumn{1}{c}{0.94} \\ 
\multicolumn{1}{c}{EfficientNet} & \multicolumn{1}{c}{Aspergillosis} & \multicolumn{1}{c}{0.93} & \multicolumn{1}{c}{0.88} & \multicolumn{1}{c}{0.98} & \multicolumn{1}{c}{0.92} \\ 
 & \multicolumn{1}{c}{Average} & \multicolumn{1}{c}{0.94} & \multicolumn{1}{c}{0.95} & \multicolumn{1}{c}{0.95} & \multicolumn{1}{c}{0.94} \\ 
\hline
 & \multicolumn{1}{c}{Normal} & \multicolumn{1}{c}{0.96} & \multicolumn{1}{c}{0.92} & \multicolumn{1}{c}{1.00} & \multicolumn{1}{c}{0.96} \\ 
\multicolumn{1}{c}{Proposed } & \multicolumn{1}{c}{Liver} & \multicolumn{1}{c}{0.99} & \multicolumn{1}{c}{1.00} & \multicolumn{1}{c}{0.98} & \multicolumn{1}{c}{0.99} \\ 
\multicolumn{1}{c}{Concatenated Model} & \multicolumn{1}{c}{Aspergillosis} & \multicolumn{1}{c}{0.98} & \multicolumn{1}{c}{1.00} & \multicolumn{1}{c}{0.95} & \multicolumn{1}{c}{0.98} \\ 
 & \multicolumn{1}{c}{Average} & \multicolumn{1}{c}{0.98} & \multicolumn{1}{c}{0.97} & \multicolumn{1}{c}{0.98} & \multicolumn{1}{c}{0.98} \\ 
\hline
\end{tabular}}
\end{center}
\end{table*}
\subsection{Proposed Work: A Pseudo code Perspective}
The pseudocode outlining the proposed concatenated AI model approach for falcon diseases classification is presented below:
\\
\hspace{2cm}
\textit{
\begin{itemize}
 \item begin
    \begin{itemize}
        \item Split the labeled falcon endoscopy dataset into two groups: training (80\%) and testing (20\%).
        \item Construct the concatenated deep learning model incorporating features from ConvNeXt and EfficientNet.
        \item Define the input (first) layer of the concatenated AI model as $128 \times 128$.      
       \item Specify the output (last) layer of the concatenated AI model to include three classes [Normal, Liver, Aspergillosis].
       \item Train the concatenated AI model using the training dataset and validate its performance.
       \item Utilize the test dataset to predict falcon endoscopy types and assess performance metrics such as accuracy, precision, recall, and F1-score.         
    \end{itemize}
\item end
\end{itemize}
}
\section{Results and Discussion}
In this study, we conducted a thorough evaluation of proposed concatenated models through extensive experiments. The assessment utilized falcon endoscopy images sourced from the Sharjah Falcon Clinic, comprising a dataset of 610 labeled images representing three distinct falcon disease classes. Falcon diseases are distributed evenly across each class, with 203 images per class. To establish robust training, testing, and validation sets, we randomly divided this dataset. The training process, conducted on the Python 3 Google Compute Engine backend of Google Colab Pro, featured GPU support and 12.7 GB System RAM. With 80\% of the dataset allocated for training and the remaining 20\% for internal validation, the training phase spanned 50 epochs, each with a batch size of five. Following an intensive training regimen, model performance underwent rigorous evaluation on the test set, while hyperparameter fine-tuning was executed using the validation set.
\subsection{Evaluation Metrics}
\label{subsec:Evaluation}
The performance of deep learning models is evaluated using metrics derived from a confusion matrix, which includes true positives (TP), false positives (FP), true negatives (TN), and false negatives (FN). Key metrics for assessing model performance are:

\textbf{Accuracy:} Measures the overall effectiveness of the model in predicting correct instances.
\begin{equation}
Accuracy = \frac{(TP + TN)}{(TP + TN + FP + FN)}     
\end{equation}
\textbf{Precision:} Assesses the model's ability to accurately identify positive instances, focusing on minimizing false positives.
\begin{equation}
Precision = \frac{TP}{(TP+ FP)}     
\end{equation}
\textbf{Recall (Sensitivity):} Evaluates the model's capacity to correctly predict positive instances, highlighting its ability to capture actual positives.
\begin{equation}
Recall = \frac{TP}{(TP+ FN)}     
\end{equation}
\textbf{Specificity:} Calculates the proportion of accurately predicted negative instances, emphasizing the model's accuracy in identifying negatives.
\begin{equation}
Specificity = \frac{TN}{(TN+ FP)}     
\end{equation}
\textbf{F1 Score:} Provides a balanced view of the model's performance by combining precision and recall, particularly useful in scenarios with imbalanced datasets.
\begin{equation}
F1 Score = \frac{2 * (Precision * Recall)}{(Precision + Recall)}    \end{equation}
These metrics collectively offer a comprehensive understanding of a model's effectiveness, strengths, and areas for improvement~\cite{zhou2021evaluating}.

\subsection{Performance of the proposed model}
The experimental results demonstrate the high accuracy and reliability of the proposed concatenated AI model solution in classifying various falcon diseases. The implemented falcon diseases classification system, utilizing the concatenated AI model, has undergone performance evaluation. The model demonstrates a notable training accuracy of 99.65\%, with a corresponding validation accuracy of 98.50\%. In terms of losses, the model achieves a training loss of 0.0865 and a validation loss of 0.0773.

The confusion matrix proves to be a potent and flexible tool in deep learning-based classification tasks, providing a comprehensive breakdown of model performance that extends beyond basic accuracy metrics. Its capacity to pinpoint specific areas for improvement and inform decision-making establishes it as a crucial element in evaluating and refining deep learning models. Presented in Figure~\ref{fig:Conf1} and ~\ref{fig:Conf2}, the confusion matrix illustrates the ConvNeXt Model, EfficientNet Model, and Concatenated ConvNeXt-EfficientNet Model's performance based on 126 test images across diverse classes. Comparative analysis against state-of-the-art models (Table~\ref{tab:Table1}) underscores the superior performance of the proposed concatenated AI model. Notably, this model demonstrates heightened prediction accuracy compared to other models examined in the study. The experimental results emphasize the potential of the proposed concatenated deep learning techniques in classifying falcon diseases, underscoring the ongoing need for research and refinement to further enhance the model's reliability and efficiency. 

ROC curve is significant in deep learning-based classification tasks as it provides a nuanced and visually interpretable performance assessment, aiding in the optimization of decision thresholds and facilitating better understanding of a model's discriminatory power across different scenarios. The area under the ROC curve serves as a quantitative measure of the model's overall discriminatory power, with higher AUC values indicating better performance. Figure~\ref{fig:ROC} show the ROC curve of the proposed concatenated AI model.

\section{Conclusion}
\label{sec:Conclusion}
\par This paper introduces an advanced concatenated AI deep learning model specifically designed for classifying falcon diseases, showcasing superior accuracy compared to existing methods. Leveraging transfer learning enhances its feature extraction and classification capabilities. The proposed model is comprehensively presented and evaluated for falcon diseases classification, with a comparative analysis against ConvNeXt and EfficientNet models. Evaluation metrics encompass accuracy, precision, recall, F1-score, and a confusion matrix. Remarkably, the concatenated AI model achieves testing accuracies of 98\%, surpassing other deep learning models considered in this study. Future efforts will focus on developing lightweight models to further improve performance and extending the model's application to broader disease analysis tasks.

\bibliographystyle{IEEEtran}

\begin{thebibliography}{00}

\bibitem{fischer2014comparison}
D.~Fischer, L.~Van~Waeyenberghe, C.~Cray, M.~Gross, E.~Usleber, F.~Pasmans, A.~Martel, and M.~Lierz, ``Comparison of diagnostic tools for the detection of aspergillosis in blood samples of experimentally infected falcons,'' \emph{Avian Diseases}, vol.~58, no.~4, pp. 587--598, 2014.

\bibitem{panthakkan2023unleashing}
A.~Panthakkan, S.~Anzar, and W.~Mansoor, ``Unleashing the power of efficientnet-convnext concatenation for brain tumor classification,'' in \emph{2023 15th Biomedical Engineering International Conference (BMEiCON)}.\hskip 1em plus 0.5em minus 0.4em\relax IEEE, 2023, pp. 1--5.

\bibitem{muller2006study}
M.~Muller, T.~Mannil, and A.~George, ``Study on the most common bacterial infections in falcons in the united arab emirates,'' in \emph{Proceedings of the 27th Annual AAV Conference, San Antonio, TX, USA}, 2006, pp. 6--10.

\bibitem{hemeiri2021retrospective}
M.~A.~A. Hemeiri, A.~A. de~la Torre, K.~Mohteshamuddin, B.~A. Degafa, and G.~Ameni, ``Retrospective study on the health problems of falcons in al ain, united arab emirates,'' \emph{bioRxiv}, pp. 2021--02, 2021.

\bibitem{alrefaei2020molecular}
A.~F. Alrefaei, ``Molecular detection and genetic characterization of trichomonas gallinae in falcons in saudi arabia,'' \emph{Plos one}, vol.~15, no.~10, p. e0241411, 2020.

\bibitem{tarello2011etiologic}
W.~Tarello, ``Etiologic agents and diseases found associated with clinical aspergillosis in falcons,'' \emph{International Journal of Microbiology}, vol. 2011, 2011.

\bibitem{muller2008outbreak}
M.~M{\"u}ller, J.~Kinne, R.~Schuster, and J.~Walochnik, ``Outbreak of microsporidiosis caused by enterocytozoon bieneusi in falcons,'' \emph{Veterinary parasitology}, vol. 152, no. 1-2, pp. 67--78, 2008.

\bibitem{mohamed2014radiographic}
M.~A. Mohamed \emph{et~al.}, ``Radiographic imaging, endoscopy and haematological profile as indispensable diagnostic tools in falcon health and disease,'' Ph.D. dissertation, Sudan University of Science and Technology, 2014.

\bibitem{sadeghi2023early}
M.~Sadeghi, A.~Banakar, S.~Minaei, M.~Orooji, A.~Shoushtari, and G.~Li, ``Early detection of avian diseases based on thermography and artificial intelligence,'' \emph{Animals}, vol.~13, no.~14, p. 2348, 2023.

\bibitem{zhou2021evaluating}
J.~Zhou, A.~H. Gandomi, F.~Chen, and A.~Holzinger, ``Evaluating the quality of machine learning explanations: A survey on methods and metrics,'' \emph{Electronics}, vol.~10, no.~5, p. 593, 2021.

\end{thebibliography}

\end{document}